\documentclass[conference]{IEEEtran}
\IEEEoverridecommandlockouts
\usepackage{graphicx} 
\usepackage[algonl,vlined,linesnumbered]{algorithm2e}

\newcommand{\EGP}{SLEGP}

\usepackage{cite}
\usepackage{amsmath,amssymb,amsfonts}
\usepackage{algorithmic}
\usepackage{graphicx}
\usepackage{textcomp}
\usepackage{xcolor}
\def\BibTeX{{\rm B\kern-.05em{\sc i\kern-.025em b}\kern-.08em
    T\kern-.1667em\lower.7ex\hbox{E}\kern-.125emX}}

\begin{document}

\title{Robotic Ad-Hoc Networks}

\author{
\IEEEauthorblockN{
Marius Silaghi, Khulud Alawaji, Mohammed Alghamdi, Akram Alghanmi, Ameerah Alsulami}
\IEEEauthorblockA{\textit{Electrical Engineering and Computer Science} \\
\textit{Florida Institute of Technology}\\
Melbourne, Florida, USA \\
msilaghi@fit.edu, \{kalawaji2017,malghamdi2018,aalghanmi2019,aalsulami2019\}@my.fit.edu}
}

\maketitle

\begin{abstract}
Practical robotic adhoc networks (RANETs), a type of mobile wireless adhoc networks (WANETs) supporting the WiFi-Direct modes common in internet of things and phone devices, is proposed based on a strategy of exploiting WiFi-Direct connection modes to overcome hardware restrictions. For a certain period of time the community was enthusiastic about the endless opportunities in fair, robust, efficient, and cheap communication created by the Adhoc mode of the WiFi IEEE 802.11 independent basic service set (IBSS) configuration that required no dedicated access points. The mode was a main enabler of wireless Adhoc networks (WANETS). This communication mode unfortunately did not get into the standard network cards present in IoT and mobile phones, likely due to the high energy consumption it exacts. Rather, such devices implement WiFi-Direct which is designed for star topologies. Several attempts were made to overcame the restriction and support WANETs, but they break at least the fairness and symmetry property, thereby reducing applicability. Here we show a solution for fair RANETs and evaluate the behavior of various strategies using simulations.

\end{abstract}

\section{Introduction}
Technologies like the Robot Operating System (ROS)~\cite{quigley2009ros} provide a powerful infrastructure once network communication is available.
Mobile Wireless area networks (MANETS) offered one of the most exciting visions for connectivity a couple of decades ago.
It was imagined how robots and other devices could use such meshes to directly exchange data when passing in proximity of each other and would be able to unleash unlimited opportunities of 
coordination~\cite{wang2003ad,maneteval15}.
However, while some other types of Adhoc networks thrived, in particular vehicular Adhoc networks (VANETS)~\cite{hasrouny2017vanet} and sensor networks, we are still in need to achieve success of mobile robotic adhoc networks (RANETs)~\cite{wang2003ad,boiten2014firechat,mozur2014off,kim2018understanding}.
One of the reasons lie in the complexity of implementing meshes using the wireless interfaces found in regular IoT devices and Android phones: a WiFi-Direct interface (WD), and a Legacy WiFi client-only interface (LC).

Some of the MANET software infrastructure has been studied and built using the IEEE 802.11 independent basic service set (IBSS) Adhoc communication mode, only present in network cards of larger devices like laptops and desktops. Attempted implementations of meshes with IoT and Android phones are mainly based on Bluetooth technology such as in Google Nearby Connections~\cite{rasmussen2019nearby,mozur2014off}.

The MANET applications have two main reasons for stumbling:

(1) On the one side, as it was also observed with applications like Firechat and Bridgefy~\cite{albrecht2021mesh,mozur2014off}, data management in such media is particularly hard due to common occurrence of large numbers of unstructured anonymous messages. 

(2) On the other side, the setup of nearby connections is challenged by the hardware restrictions in IEEE WiFi-Direct.  The data management problem was addressed in many previous publications~\cite{dhannoon2013content}, and in this work we focus on this second problem.

The common solution to the Adhoc connectivity problem of IEEE WiFi-Direct has been to use Bluetooth for communication, while broadcasting MAC addresses using WiFi-Direct broadcast beacons~\cite{rasmussen2019nearby,pradeep2022moby,burton2019secure}. The main weakness of this approach is that the range of Bluetooth is only 10 meters and cannot support significant exchanges when  participants are casually passing by each other.
The main reason for not using WiFi-Direct itself for Adhoc communication is that its design was solely for star topologies, with a Group Owner (GO) in the center that forwards messages between its Group Member (GM) clients. Any hop-based message forwarding by another participant needs some type of trick.

Previous efforts tried to come up with topologies where Group Owners or WiFi-Direct group members (GM) use their Legacy WiFi endpoint (LC) to join other groups and route packets, thus extending the range of the network~\cite{teofilo2015group,casetti2015demonstration}. However, setup of these topologies is complex in dynamic settings and it is even harder to maintain them
in the context of moving participants.

In this work we propose a different strategy, based solely on WiFi but based on continuously
alternating between GO group owner and GM group member modes for the WiFi-Direct interface.

\section{Background}
Mobile robots, IoT devices, or Android phones have multiple radio communication technologies: BlueTooth, Legacy WiFi client (LC), WiFi-Direct (WD), near field communication (NFC). NFC supports communication up to distances of 4cm.
The BlueTooth radio has a range of 10m.
The wireless interfaces of mobile  WiFi-Direct, just like Legacy WiFi clients, have a range of 200m. The WiFi-Direct standard was originally introduced under the name of WiFi P2P. However, the intended architecture supported is a star topology where the device in the center is called Group Owner (GO) while the rest of the participants in the group are sometimes referred to as Group Members (GM). The radio signals in WiFi-Direct are similar to the ones in Legacy WiFi, but the setup differs, introducing a negotiation phase for the selection of the Group Owner. This negotiation can be performed according to one of three supported methods: (1) standard: where participants engage in a distributed coin throwing with preferences~\cite{maraj2019learning}, (2) autonomous: where a device is pre-configured as group owner, and (3) persistent: where the participants remember their roles and just resume them.
The Legacy WiFi client (LC) conforms with the IEEE 802.11 standard but does not support the useful but energy hungry adhoc mode.

\begin{figure}[!h]
\includegraphics[width=3in]{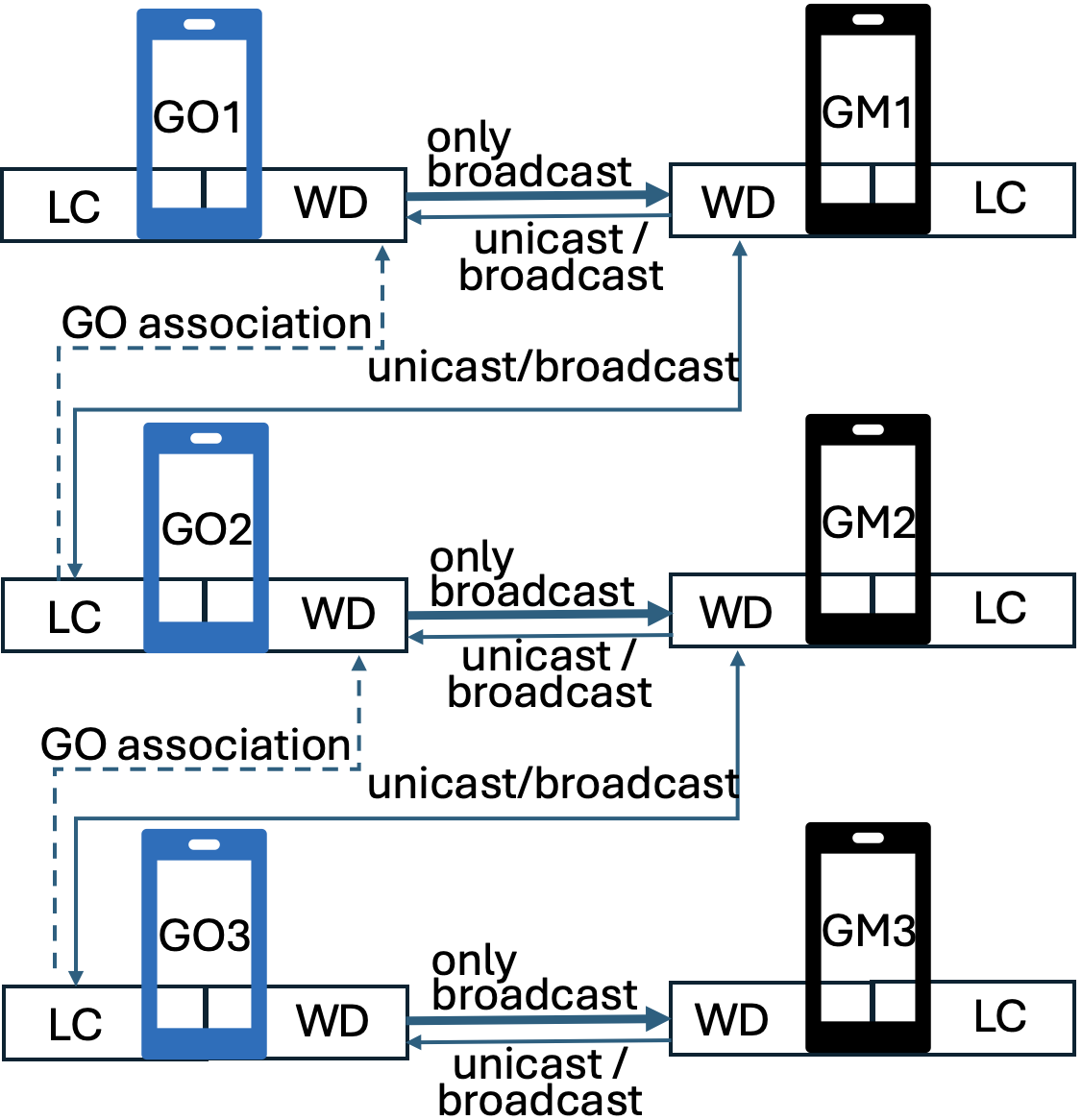}
\caption{RANET with broadcast segments}\label{fig:manetcasetti}
\end{figure}

While reasonable success was reached in physical building of mobile AdHoc networks using short distance BlueTooth connections~\cite{boiten2014firechat,albrecht2021mesh}, several researchers have addressed the challenge of exploiting the longer range WiFi signals.
One of the first solutions is proposed in~\cite{casetti2015demonstration}.
The authors observed that group owners cannot communicate with each other by unicast since their IP addresses are all hard-codded by systems like Android to the same value, namely {\tt 192.168.49.1}.
Further, when GOs use their 
Legacy WiFi client interface to join as members of other groups, their routing tables can no longer unicast messages to their own group members, but can still communicate to these ones by UDP broadcast.
Their full connectivity solution is depicted in Figure~\ref{fig:manetcasetti}.

In the solution in~\cite{duan2014wi,casetti2015demonstration,casetti2017data}, bidirectional communication is only possible with UDP broadcast on the segments from each GO to its group members, and such a segment must be present between each two groups, at least in one direction.

\begin{figure}[!h]
\includegraphics[width=3in]{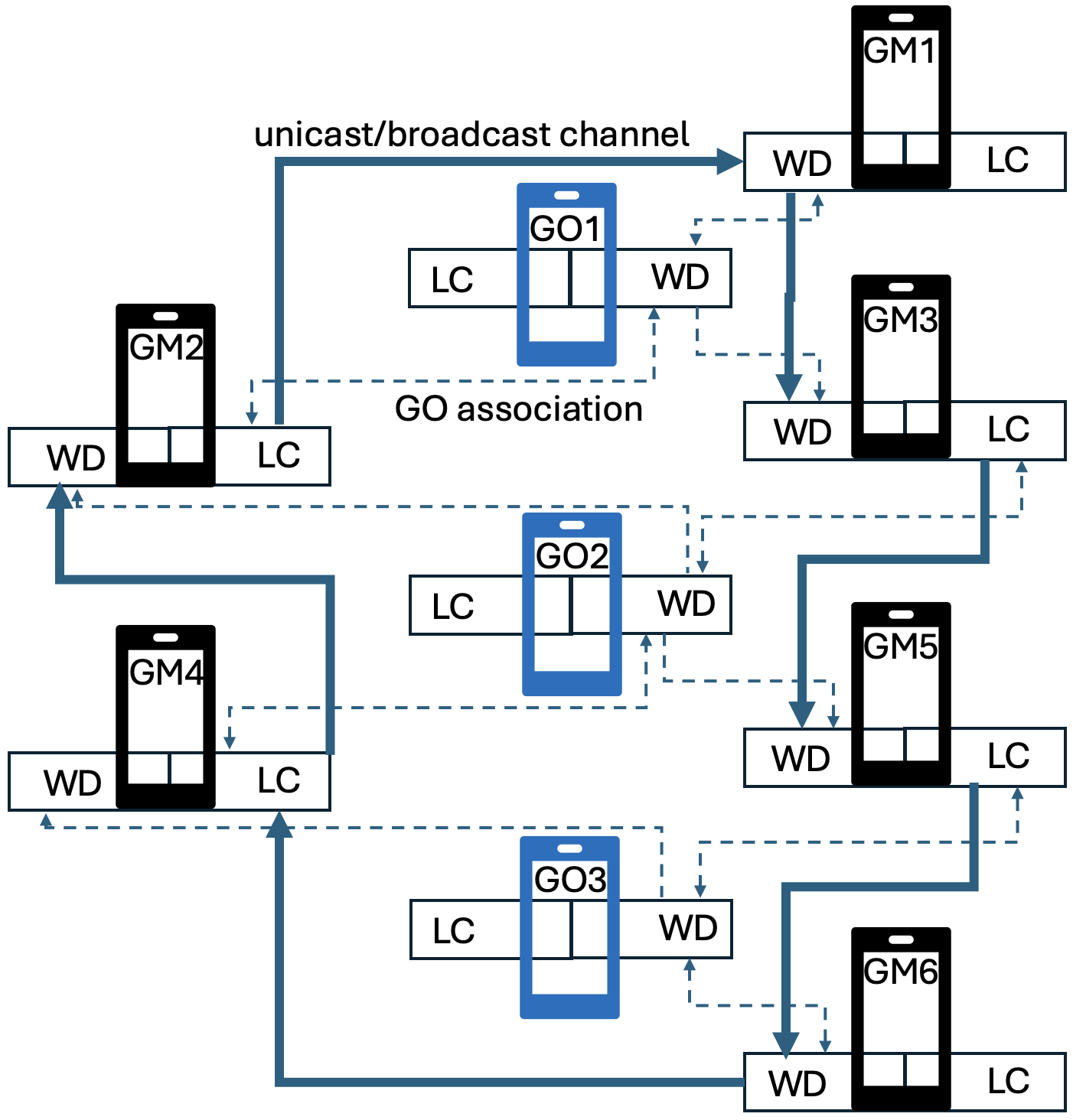}
\caption{RANET with two client relays per connection}\label{fig:manetteofilo}
\end{figure}

A family of alternative solutions is proposed in~\cite{teofilo2015group,funai2017enabling,shahin2015efficient}, based on an architecture where each two group owners are connected via two clients used as relays, each of them attached to a GO with the WD interface and attached to the other GO using the LC Legacy WiFi client interface, and specialized in forwarding the data from the WD connected GO to the GO on the LC Legacy WiFi client interface. The obtained mesh architecture is shown in Figure~\ref{fig:manetteofilo}.

Another family of approaches is based on the adoption of separate access points (APs), mobile devices that would allow 
client devices
and WiFi-Direct group owners to seamlessly communicate to each other using the original design of the standard~\cite{xu2017wi}. However, this solution requires users to carry such access points devices separately, causing discomfort.

\section{Proposed Short-Lived Ephemeral Groups Protocol}

The main challenge with existing solutions for content delivery like the ones in Figures~\ref{fig:manetcasetti} and~\ref{fig:manetteofilo}, is that the connection pattern is hard to maintain in the highly dynamic environments of naturally-moving
robors and humans.
Setups protocols, like the Smart Group Formation~\cite{casetti2017data} and Efficient Multi-group formation and Communication (EMC)~\cite{shahin2015efficient}, typically detect nearby groups with members and group owners, before deciding whom and how to contact. 

Our proposal is to bypass all the setup of the complex structures in the previously proposed architectures, and rather to dynamically create ephemeral groups which will be disbanded after relatively short sessions of exchanges, creating opportunities to connect new devices and disseminate wider the available messages.

The Short-Lived Ephemeral Groups Protocol (\EGP{}) is described in Algorithm~\ref{alg:egp}. 
At each moment, the WiFi-Direct interfaces of the devices may be either in group owner (GO) mode, or in group member (GM) state. The searching state, when a device wants to become a Group Member but is still in search of a group owner, is also referred to as part of the GM mode. The devices will switch between these states randomly (see Lines~\ref{ln:go} and~\ref{ln:gm}), but only after a minimum duration elapses in the 
current state. The minimum duration is based on
the corresponding state, with recommended ratios that depend on the density of devices.

For example, if a device senses a number of $N$ other reachable devices around, it may decide a duration in GM that is $\frac{N-1}{c}$ times larger than in a GO state, to heuristically support $c$ groups. Also, the duration in GO state should not be encouraged to surpass the one in GM state, to avoid the generation of many disjoint groups.
Moreover, users may want to strongly reduce the GO time, and even suppress it completely, in order to reduce power consumption for their devices. 

The selection of the GO can be based on the signal strength and direction of movement. The selection from the list of availabilities, if based on a probability distribution that favors 
GOs that are further apart but found in a relative  movement that is bringing it closer,  potentially enables longer connection time. Prior to connection, distance is commonly evaluated based on signal strength, but after connection it can be updated by GPS location information shared by group devices.

A receiver thread in the protocol at Line~\ref{ln:receiver} is in charge of receiving messages and integrating them in 
the local database. The receiver thread and sender threads are assigned relative priorities that are enforced by the \EGP{} protocol scheduler.

\begin{algorithm}
\caption{
Ephemeral Groups Protocol (\EGP{}) for gossip: inner loop behaviors. The receiver/sender behaviors have relative priorities enforced by the protocol scheduler.
}\label{alg:egp}
  \DontPrintSemicolon
  \SetAlgoLined
  \SetKwFor{ForEach}{ForEach}{}{}
  \SetKwFunction{receiver}{receiver}
  \SetKwFunction{onGO}{GO-main-loop}
  \SetKwFunction{onGM}{GM\&Legacy-main-loop}
  \SetKwProg{Pn}{proc}{:}{\KwRet}
  \SetKwProg{proc}{proc}{:}{}
\lnl{ln:GO}  \proc{\onGO{}}{
        update GM based on reachability\;
        $message$ := select message to be broadcast\;
        broadcast $message$ to current GMs\;
        \If{time as GO > minimum GO service time}{
\lnl{ln:go}            random switch to GM\; 
        }
  }
\lnl{ln:GM}  \proc{\onGM{}}{
        update GO based on reachability\;
        \If{have GO} {
            $message$ := select message to be broadcast\;
            broadcast $message$ to current GMs\;
        }
        \If{time as GM > minimum GM service time}{
\lnl{ln:gm}            random switch to GO\; 
        }
  }

\lnl{ln:receiver}  \proc{\receiver{}} {
  \ForEach{received $message$} {
   integrate message in local database\;
   update queue of messages to broadcast\;
  }
  }
\end{algorithm}

When found in GM mode, a device schedules the competing procedures at Lines~\ref{ln:GM} and~\ref{ln:receiver}, based on their priorities, where the receiver procedure only works when there are messages available in the incoming queue.

When found in GO mode, a device schedules the competing procedures at Lines~\ref{ln:GO},~\ref{ln:GM} and~\ref{ln:receiver}, based on their priorities. As before, the receiver procedure only works if there are messages available in the incoming queue. Additionally the \onGM{} procedure is only selected in GO mode if the Legacy WiFi interface is managed by \EGP{}, which happens if it is associated to an \EGP{} supporting AP, or manually assigned by user to the local \EGP{} protocol manager.

\begin{figure}[!h]
\includegraphics[width=3.2in]{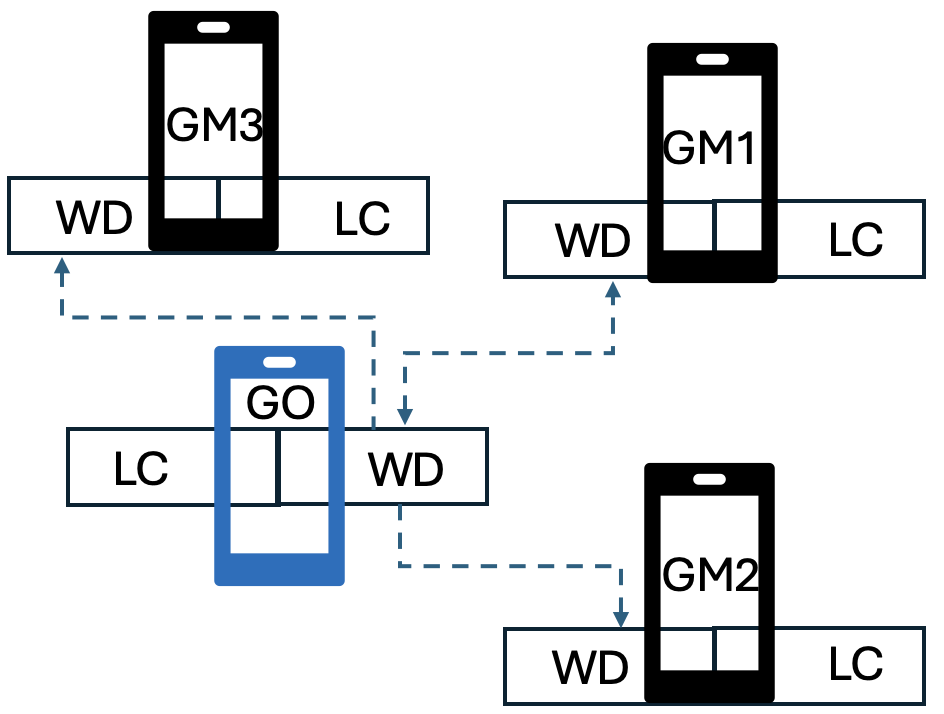}
\caption{Simple ephemeral \EGP{} cell. Legacy WiFi client interfaces are only used optionally for additional connections.}\label{fig:egp}
\end{figure}

In order to decide what messages to select for broadcasting to maximize the utility of the dissemination process, we inspire from the utility based dissemination in vehicular adhoc networks, VANETS~\cite{dhannoon2013content}.

In this model, we assume that a number of $A$ new devices are approaching the current device with average contact time $T_A$, while $B$ new devices are traveling together with the current device in the same direction with average contact time $T_B$. 
Multiple message queues may be prepared for transmission, each queue including messages of a certain utility.

A queue of messages of size $B_s$ is extracted from the database at 
time periods $P_{reload}$, each message in this queue having utility $u_M$ and transmission time based on message size, $v_M$. In this scenario,
the utility of sending messages from this queue is:

\begin{equation}
    \frac{\delta U}{\delta t} = u_M\left(\frac{A\cdot B_s}{P_{reload}}+B\cdot v_M\right)
\end{equation}

Different queues are used to sort messages by their utility, and are served by schedules that maximize the total utility.
The number of the messages most important for the sender may be much smaller than the number of other messages.

The utility of the messages changes dynamically, decreasing as the reachable peers learn them, and increasing when new peer devices appear.

\section{Experimentation}

Compared with wider area static meshes developed earlier, the advantage of the proposed technique is in robustness to dynamic and adverse environments, since reconfiguration is intrinsic.
While the maximum throughput achievable with our proposal theoretically must be smaller than the one achievable with 
a static mesh connected with one of the corresponding techniques, like~\cite{casetti2017data}, our simulations still show that it is reasonable. Static meshes do not waste time for reconfiguration.

However, if the devices are moving, and the reachability graph changes very dynamically, configurations like the ones in Figures~\ref{fig:manetcasetti} and~\ref{fig:manetteofilo} are harder to be maintained. Even if discovery would be used to attempt their dynamic reconstruction, they enable disconnected graphs that would yield unexpected and unintuitive lack of data sharing between participants that spend much time together.

When the Legacy WiFi are allowed to connect to other GOs, architectures like the ones in Figures~\ref{fig:manetcasetti} and~\ref{fig:manetteofilo} will nevertheless be spontaneously obtained and exploited, as long as the main transport is UDP broadcast. However, in our experiments we assume that Legacy WiFi connections are reserved by users for Internet access using their preferred APs, and will not be available to the \EGP{} protocol manager.

In order to evaluate the throughput achievable with the \EGP{} protocol,
we use a benchmark simulating an urban downtown circuit of length $L$, $L=1 km$, where a number of $n$ devices travel clockwise and another $n$ devices travel counterclockwise with a velocity of $1 m/s$ and spaced at $D=20m$ from each other.

Each device starts having a number of $M$ messages, generated by itself, with high utility for itself.
The devices collaborate in disseminating to each other messages and 
try to maximize their own utility by giving a preference to the dissemination of their own personal messages.

Selection of messages to send next prefers personal messages which are allocated at least 50\% of the bandwidth, but not less that the priority of other messages.
Messages sent are marked. If a message is selected for transmission and found marked as already transmitted recently, then the mark is cleared and the message is not sent at this time. 
When a GM is searching for GOs and finds multiple reachable ones, while the selection could use described heuristics to maximize time of pairing, in current experiments the selection is uniformly random among choices.

\begin{figure}[!h]
\includegraphics[width=3.2in]{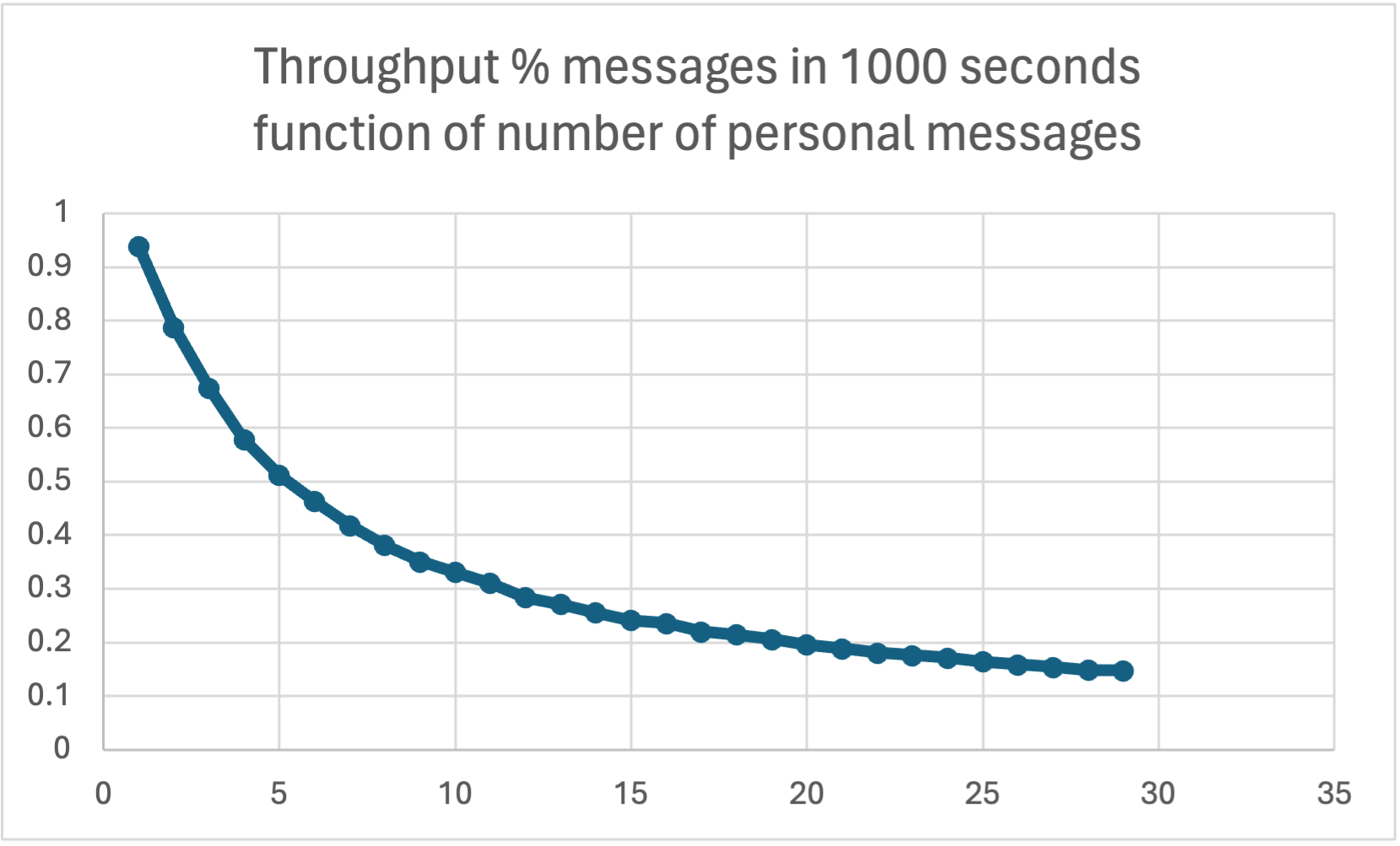}
\caption{Throughput as \% of the total number of possible delivered messages in the system, given the number of personal messages championed by each user.}\label{fig:egp1}
\end{figure}

\begin{figure}[!h]
\includegraphics[width=3.2in]{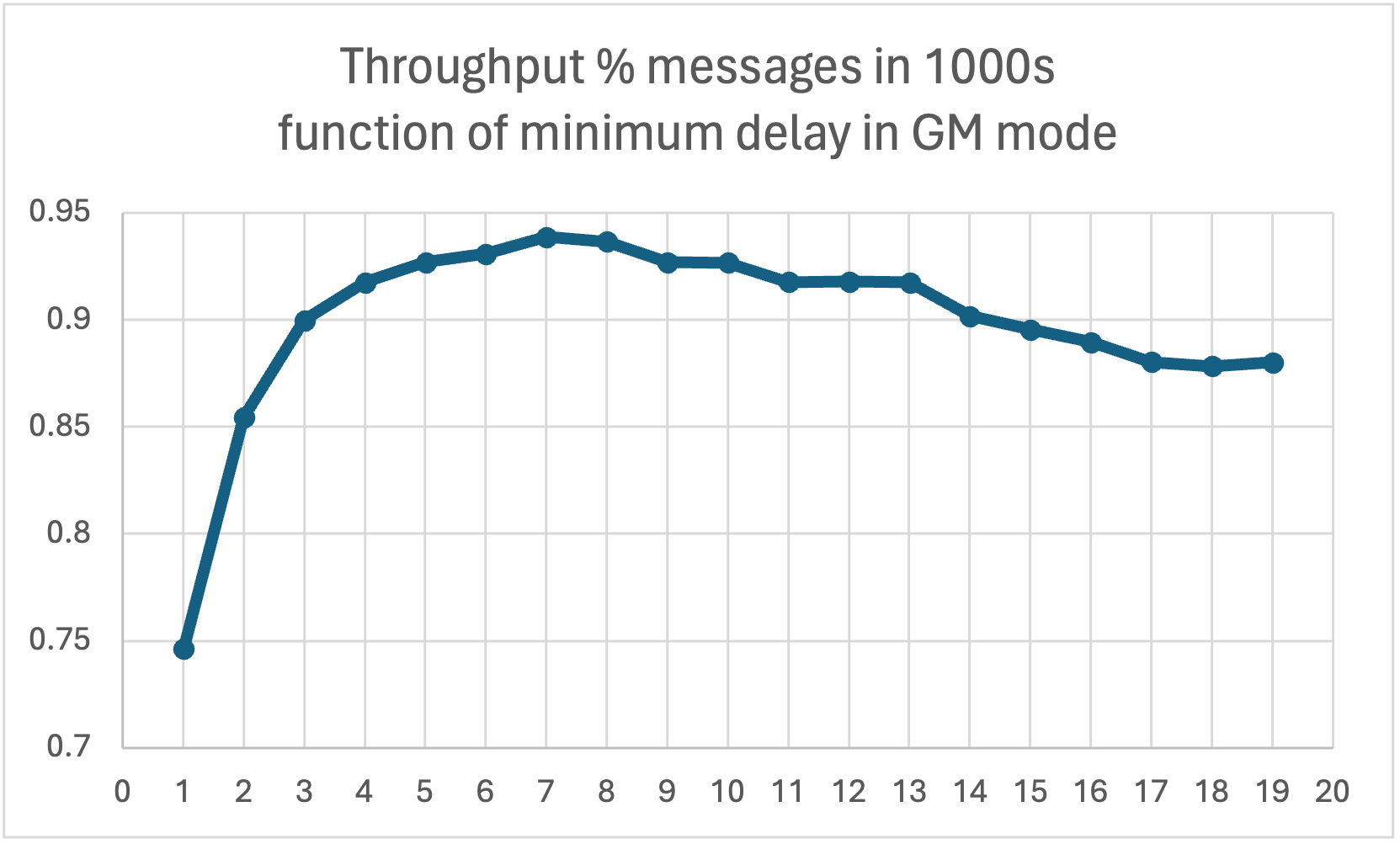}
\caption{Throughput as \% of the total number of possible delivered messages in the system, given the minimum number of seconds spent in GM mode before switching, when the GO min time is 9s.}\label{fig:egp2}
\end{figure}

\begin{figure}[!h]
\includegraphics[width=3.2in]{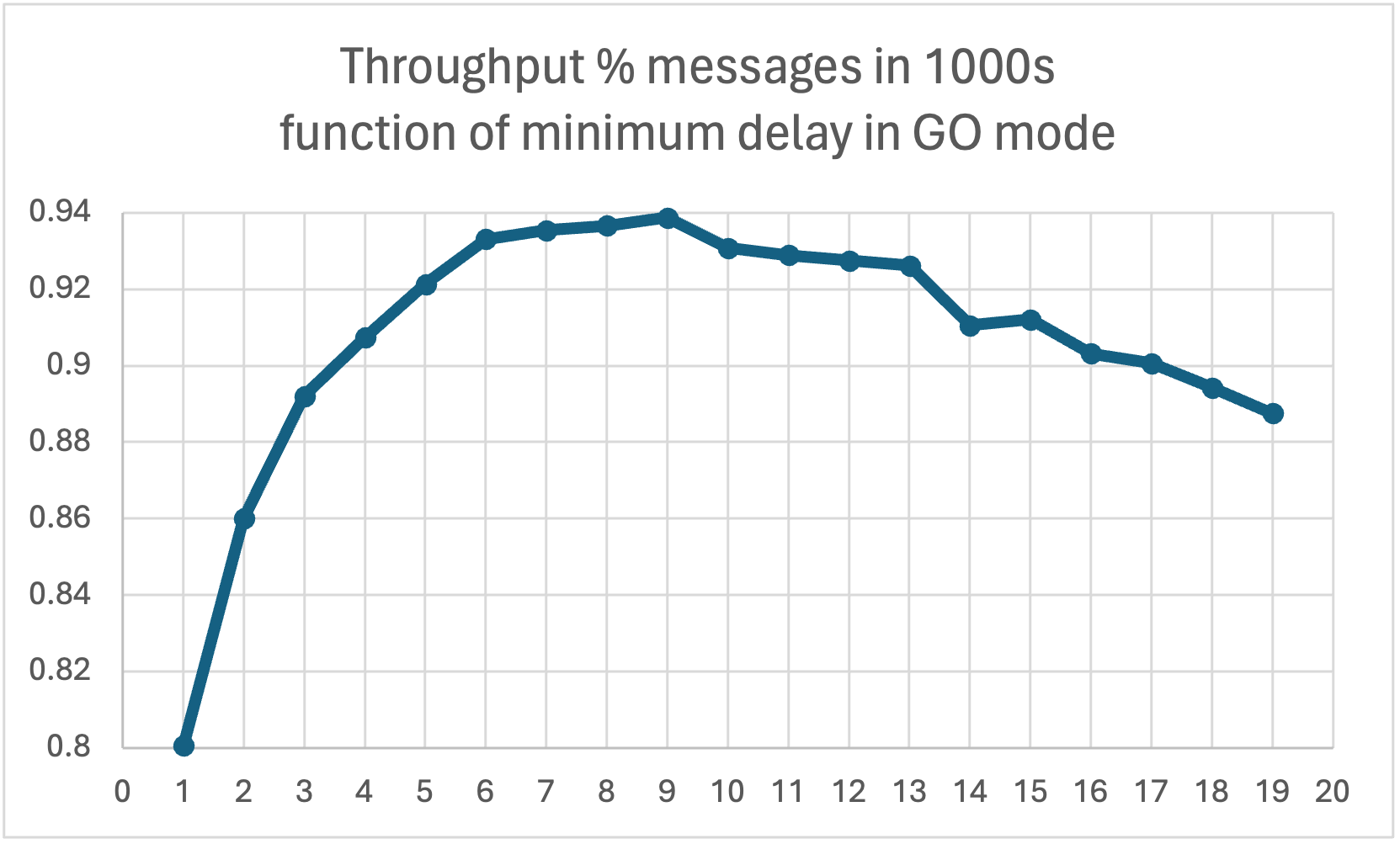}
\caption{Throughput as \% of the total number of possible delivered messages in the system, given the minimum number of seconds spent in GO mode before switching, when the GM min time is 7s.}\label{fig:egp3}
\end{figure}

\begin{figure}[!h]
\includegraphics[width=3.2in]{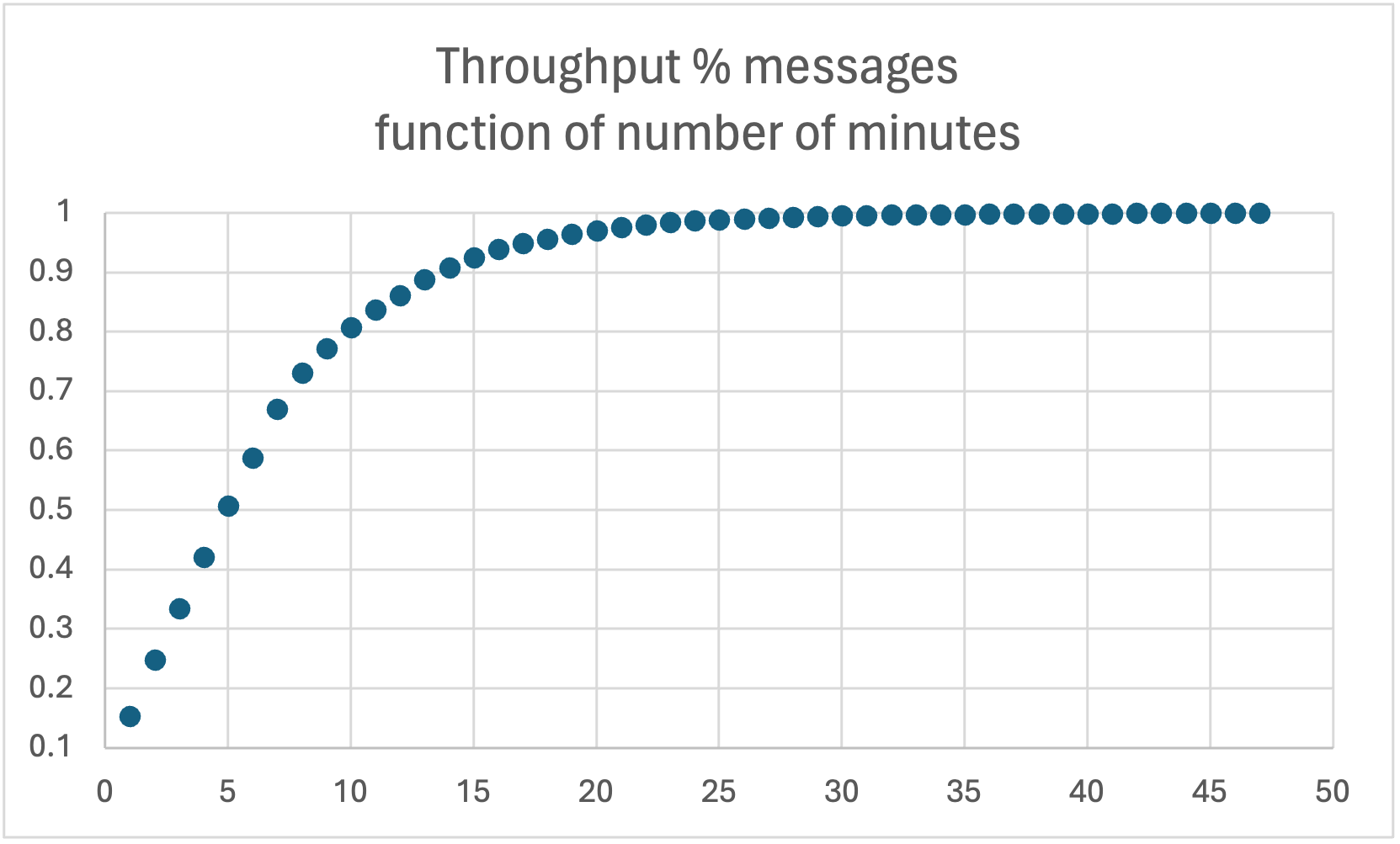}
\caption{Throughput as \% of the total number of possible delivered messages in the system, given the number of simulated minutes of interaction. 100\% delivery was achieved after 47 simulated minutes, with min 9s in GO mode, min 7s in GM mode and 1 personal message per participant.}\label{fig:egp4}
\end{figure}

The experimentation with the simulated circuits is used to evaluate strategies based on the amount of exchanged messages after a certain time of interaction.
The amount of communication is represented by the throughput. The {\bf throughput} for us is the fraction of messages delivered in $T$ seconds out of the total number of messages deliveries that can be occur to all participant devices.
With $2n$ devices and $M$ messages generated by each  device, the total number of unique messages in the system is $2nM$, and the total number of new message deliveries is $(4n^2-2n)M$.

Our simulated experiments were run with 
a number of devices $n=50$, spaced at $20m$ in each direction. 
After the minimum amount in each mode, the mode is switched every second with 50\% probability.
When the throughput is computed after $T=1000$ seconds when all devices travel on the circuits,
Figure~\ref{fig:egp1} shows that throughput decreases slowly with the number of unique messages in the system.

Also computed after $T=1000$ seconds of exchanges, the throughput as function of minimum duration in GM mode and in GO mode has a local maximum at values $7s$ in GM mode and $9s$ in GO mode (see Figures~\ref{fig:egp2} and~\ref{fig:egp3}). 

As function of number of minutes $T$ in which the throughput is computed, the progression is depicted in Figure~\ref{fig:egp4} for the case of 1 message generated per device. It can be observed that within 47 seconds, all the messages are delivered to all peers, achieving a 100\% throughput. A throughput of 90\% is achieved in less than 15 minutes.

The evaluation is not based on utility as in~\cite{dhannoon2013content}, but on number of messages, under the simplifying assumption that all messages are equally useful to everyone, but extension to evaluation for variable utility is possible in future work.

\section{Conclusions}

A new protocol was proposed for transforming WiFi-Direct 
devices
into a gossiping wireless Mobile Robot Adhoc Networks based on the formation of short-lived ephemeral groups. These groups
keep being formed and disintegrating when the members switch modes to new regroup at random periods of time. The randomness is intended in order to break patterns that would lock pairs of devices to the same groups or apart from each other. Non-random switching would lock a pair of devices to always be GOs in the same time, and therefore never pair with each other even when travelling together.

The Short-Lived Ephemeral Groups Protocol (EGP) sees devices switch between GO and GM modes for quantas of time distributed around means depending on the detected density of devices. Simulations with hundreds of devices traveling in both directions on a closed circuit and initialized with parameters compatible with ones measured experimentally on hardware in~\cite{camps2013device}, suggest that switching times of $9$ seconds would maximize the throughput measured in terms of total messages delivered.
The protocol is useful to support content broadcasting such a news and general interest data in situations of Internet disruptions.

\bibliographystyle{IEEEtran}
\bibliography{wifi.bib,wifidirect.bib}

\end{document}